\pdfoutput=1

\documentclass[11pt]{article}
\usepackage{graphicx}
\usepackage{acl} 
\usepackage{CJK}

\usepackage{multirow}
\usepackage{times}
\usepackage{latexsym}
\usepackage{booktabs}
\usepackage{url}
\usepackage[T1]{fontenc}

\usepackage[utf8]{inputenc}

\usepackage{microtype}
\usepackage{tikz}
\usetikzlibrary{shapes.geometric}
\newcommand{\warningsign}{\tikz[baseline=-.75ex] \node[shape=regular polygon, regular polygon sides=3, inner sep=0pt, draw, thick] {\textbf{!}};}

\usepackage{amsthm}
\theoremstyle{application}

\title{Huatuo-26M, a Large-scale Chinese Medical QA Dataset}

\author{Jianquan Li$^{\beta*}$, Xidong Wang$^{\alpha,\beta}$\thanks{The first two authors contributed to this paper equally}~, Xiangbo Wu$^\beta$, Zhiyi Zhang$^\beta$, Xiaolong Xu$^\beta$, \\ \textbf{Jie Fu$^\gamma$, Xiang Wan$^\alpha$, Benyou Wang$^{\alpha,\beta\includegraphics[width=0.45cm]{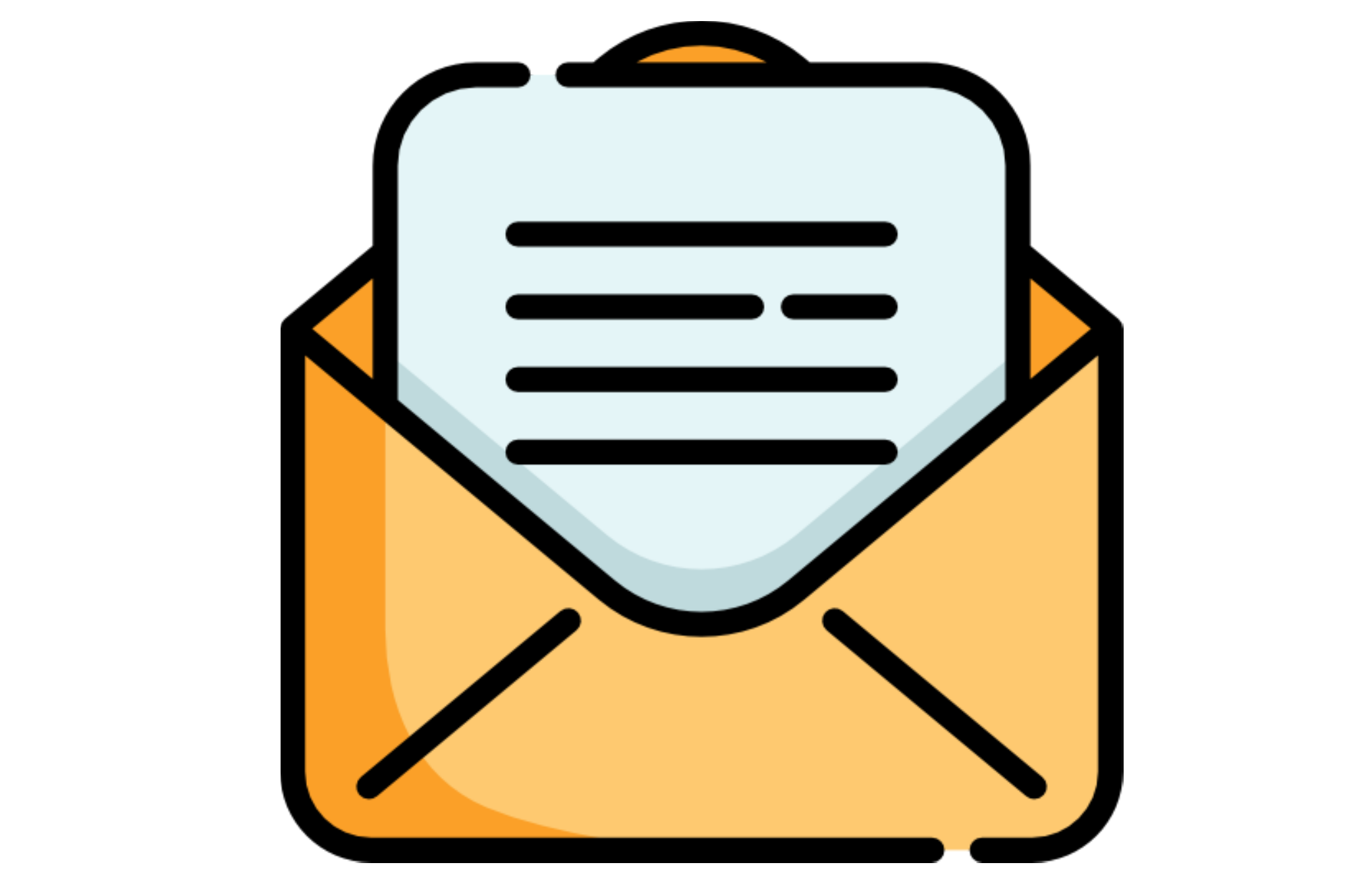}}$} \\
$^\alpha$ Shenzhen Research Institute of Big Data\\
$^\beta$ The Chinese University of Hong Kong, Shenzhen \\
$^\gamma$ Beijing Academy of Artificial Intelligence \\
\includegraphics[width=0.45cm]{figures/email.pdf}\texttt{wangbenyou@cuhk.edu.cn} 
}

\begin{document}
\maketitle
\begin{abstract}
In this paper, we release a \textbf{largest} ever medical Question Answering (QA) dataset with \textbf{26 Million} QA pairs.  We benchmark many existing approaches in our dataset in terms of both retrieval and generation. Experimental results show that the existing models perform far lower than expected and the released dataset is still challenging in the pre-trained language model era. Moreover, we also experimentally show the benefit of the proposed dataset in many aspects:  (i) trained models for other QA datasets in a zero-shot fashion; and (ii) as external knowledge for retrieval-augmented generation (RAG); and (iii) improving existing pre-trained language models by using the QA pairs as a pre-training corpus in continued training manner.  We believe that this dataset will not only contribute to medical research but also facilitate both the patients and clinical doctors. See \url{https://github.com/FreedomIntelligence/Huatuo-26M}.
\end{abstract}

\section{Introduction}

Pre-trained language models have made great progress in natural language processing (NLP) and largely improve natural language understanding and natural language generation. 
This inspires researchers to apply PLMs for fields that were not considered the core playground of NLP, for example, medicine. However, the first \textit{bottleneck} for medicine using PLMs is the \textit{data}, like most other breakthroughs in artificial intelligence that starts with data collection.

To break the bottleneck, this work collects the largest medical Chinese QA dataset that also might enhance medical research. Note that there are 1.4B population speaking Chinese as their native language, and more importantly, the medical care for them (particularly the mainland of China) is generally far below the western counterpart (e.g., English-speaking and developed countries) 
\footnote{see  \url{https://en.wikipedia.org/wiki/World_Health_Organization_ranking_of_health_systems_in_2000}}.

\paragraph{Dataset}
We collect the largest medical QA dataset from various sources as below:  (i) collected from an online medical consultation website; (ii) automatically extracted from medical encyclopedias, and (iii) automatically extracted from medical knowledge bases. 
After text cleaning and data deduplication, we obtained the largest Chinese medical QA dataset, containing \textbf{26 Million} QA pairs.
We call this dataset `Huatuo-26M' to commemorate the great Chinese physician named Hua Tuo, who lived around 200 AC. As seen from Table~\ref{tab:datasets} that this work has expanded the existing medical domain QA dataset by more than two orders of magnitude, even larger than most QA datasets in the general domain.

\paragraph{Benchmark} 
Based on the collected dataset, we benchmark classical methods in the field of retrieval: For sparse retrieval, we test the performance of BM25~\cite{robertson2009probabilistic} and DeepCT~\cite{dai2019context}, and for dense retrieval, we test the performance of DPR~\cite{karpukhin2020dense}. At the same time, we also trained some of the auto-regressive language models namely GPT2~\cite{brown2020language} and T5~\cite{raffel2020exploring}. The results suggest the task is still challenging using pre-trained language models, probably because the medical domain involves more expert knowledge than the general domain. 

To further show the usefulness of the collected dataset, we leverage the collected dataset in three use cases:  (i) transfer to other QA datasets; (ii) as external knowledge for RAG; and (iii) as a pre-trained corpus.


\begin{table*}[]\footnotesize
\addtolength\tabcolsep{-3pt} 
    \centering
    \begin{tabular}{@{}llllc@{}}
    \toprule
    Dataset& Lang    & Domain  & Source    & \#Q  \\ \midrule
    MedHop~\cite{welbl2018constructing} & English & Medical & MEDLINE   & 2.5K \\
    BiQA~\cite{lamurias2020generating} & English & Medical & Online Medical forum & 7.4K \\
    HealthQA~\cite{zhu2019hierarchical} & English & Medical & Medical-services website & 7.5K \\
    MASH-QA~\cite{zhu2020question} & English & Medical & Medical article website & 35K \\
    MedQuAD~\cite{ben2019question} & English & Medical & U.S. National Institutes of Health (NIH) &  47K\\
    ChiMed~\cite{tian2019chimed} & Chinese & Medical & Online Medical forum & 47K \\
    MedQA~\cite{jin2020disease}  & EN\&CH  & Medical & Medical Exam& 60K  \\
    webMedQA~\cite{he2019applying} & Chinese & Medical &  Medical consultancy websites & 63K\\
    CliCR~\cite{vsuster2018clicr}  & English & Medical & Clinical case reports& 100K \\ 
    cMedQA2~\cite{8548603} & Chinese & Medical & Online Medical forum & 108K \\ 
    \hline
    \textbf{Huatuo-26M} & \textbf{Chinese} & \textbf{Medical} & \textbf{Consultation records, Encyclopedia, KBs} & \textbf{26M}\\
    \midrule
    TriviaQA~\cite{joshi2017triviaqa} & English & General & Trivia    & 96K \\
    HotpotQA~\cite{yang2018hotpotqa} & English & General & Wikipedia & 113K \\
    SQuAD~\cite{rajpurkar2016squad}  & English & General & Wikipedia   & 158K \\
    DuReader~\cite{he2017dureader} & Chinese & General & Web search& 200K \\
    Natural Questions~\cite{kwiatkowski2019natural} & English & General & Wikipedia & 323K\\
    MS MARCO~\cite{nguyen2016ms} & English & General & Web search& 1.0M \\
    CNN/Daily Mail~\cite{see2017get}    & English & General & News      & 1.3M \\
    PAQ~\cite{lewis2021paq}	     & English	& General & Wikipedia	& 65M \\
   \bottomrule
    \end{tabular}
    \caption{Existing QA dataset.  }
    \label{tab:datasets}
\end{table*}

\paragraph{Use case I: Transfer for other QA dataset}
Since the Huatuo-26M dataset is large, we also expect that the models trained by the dataset could encapsulate general medical knowledge. Therefore, we use the trained models on two existing medical QA datasets, namely cMedQA2~\cite{8548603} and webMedQA~\cite{he2019applying}. Experimental results show that the model can achieve competitive performance even in few or zero samples.

\paragraph{Use case II: As an external knowledge for RAG} 
Large-scale medical QA datasets themselves explicitly contain rich medical knowledge, and we leverage it as external knowledge in the context of retrieval-augmented generation~\cite{lewis2020retrieval}.  Experimental results on cMedQA2 and webMedQA datasets show that using this dataset as an external knowledge base can greatly improve the quality of generated texts.

\paragraph{Use case III: As a pre-trained corpus}

Considering that the scale of the data set is comparable to that of pre-training corpora of general pre-trained language models,
we use the text corpus of Huatuo-26M as a pre-trained corpus that could inject implicit knowledge into the model through pre-training. We improve Bert and RoBERTa in a continuously-training manner on the dataset by using QA pairs as pre-training corpora. The experimental results show the performance of pre-trained models on biomedical tasks could be largely improved by using Huatuo-26M as an additional pre-training corpus.

\paragraph{Contributions} of this work are as follows: (\textbf{i}) We release the largest Chinese Medical QA dataset (with \textbf{26,504,088} QA pairs);
(\textbf{ii}) we benchmark some existing models for the proposed methods for both retrieval and generation;
and (\textbf{iii}) we explore some additional usage of our dataset, for example,   transfer for other QA datasets,  train as external knowledge for RAG, and train as a pre-trained corpus.


\section{Dataset Creation}
We have collected a variety of medical knowledge texts from various sources and unified them in the form of medical question-and-answer pairs. The main resources include an online medical QA website,  medical encyclopedias, and medical knowledge bases. See Appendix \ref{appendix:dataset_example} for specific examples from different sources. Here we will introduce the details of data collection from the above three data sources.



\subsection{Online Medical Consultation Records}

\paragraph{Data Sources}
We collected data from a website for medical consultation \footnote{Qianwen Health in https://51zyzy.com/},  consisting of many online consultation records by medical experts. Each record is a QA pair: a patient raises a question and a medical doctor answers the question. The basic information of doctors (including name, hospital organization, and department) was recorded.

\paragraph{Data Processing}
We directly crawl patients' questions and doctor's answers as QA pairs, getting 31,677,604 pairs. 
Subsequently, we removed the QA pairs containing special characters and removed the repeated pairs. Finally, we got 25,341,578 QA pairs.





\subsection{Online Medical Encyclopedia}
\paragraph{Data Sources}
We extract medical QA pairs from plain texts (e.g., medical encyclopedias and medical articles). We collected 8,699 encyclopedia entries for diseases and 2,736 encyclopedia entries for medicines on  Chinese Wikipedia~\footnote{zh.wikipedia.org/wiki/}. Moreover, we crawled 226,432 high-quality medical articles from the Qianwen Health website\footnote{https:/51zyzy.com/}.


\paragraph{Data Processing}
We first structure an article. Each article will be divided into title-paragraph pairs. For example, such titles in articles about medicines could be usage, contraindications, and nutrition; for articles about medicines about diseases, they could be diagnosis, clinical features, and treatment methods. We remove the titles of paragraphs that have appeared less than five times, finally resulting in 733 unique titles. Based on these titles, we artificially design templates to transform each title into a question that could be answered by the corresponding paragraph. Note that a disease name or a drug name could be a placeholder in the templates.  See the appendix \ref{appendix:wiki_bases_templates} for details.



\subsection{Online Medical Knowledge Bases}

\paragraph{Data Sources}
Some existing knowledge bases explicitly store well-organized knowledge, from which we extract medical QA pairs. We collect data from the following three medical knowledge bases: \textbf{CPubMed-KG}~\footnote{https://cpubmed.openi.org.cn/graph/wiki} is a knowledge graph for Chinese medical literature, which is  based on the large-scale medical literature data from the Chinese Medical Association;  
\textbf{39Health-KG}\footnote{https://github.com/zhihao-chen/QASystemOnMedicalGraph} and
\textbf{Xywy-KG}\footnote{https://github.com/baiyang2464/chatbot-base-on-Knowledge-Graph} are  two open source knowledge graphs. See basic information is shown in Tab.\ref{tab:Kb}.

\begin{table}[h]
    \footnotesize
    \addtolength\tabcolsep{-3pt} 
    \centering
    \begin{tabular}{lrrrrrr}
    \toprule
      & \# entity type & \#relation & \#entity & \#triplets \\ \midrule
    CPubMed-KG   & - & 40 & 1.7M & 4.4M \\
    39Health-KG   & 7 & 6  & 36.8K & 210.0K\\ 
    Xywy-KG   & 7  & 10 &   44.1K & 294.1K     \\
 \bottomrule
    \end{tabular}
    \caption{Basic statistics of the three knowledge bases.  }
    \label{tab:Kb}
\end{table}


\paragraph{Data Processing}
We clean the three knowledge graphs by removing invalid characters and then merge entities and relationships among entities among these three knowledge graphs,  resulting in 43 categories. Each category is associated with either a relationship between entities or an attribute of entities.  Subsequently, we manually design templates to convent each category to a  \textit{question}. The \textit{question}  is either 1) querying the object entity based on the subject entity or 2) querying an attribute of an entity. The object entity will be the \textit{answer} w.r.t the \textit{question} in both cases.
Finally, we obtained 798,444 QA pairs by constructing questions and answers with corresponding templates. See the appendix \ref{appendix:knowledge_bases_templates} for details.





\begin{table}[]
    \footnotesize
    \centering
    \begin{tabular}{@{}lrcc@{}}
    \toprule
    Composition     & \# Pairs & Len(Q) & Len(A) \\ \midrule
    Huatuo-26M Train   & 26,239,047 & 44.6 & 120.7 \\
    Huatuo-26M Test   & 265,041  & 44.6 & 120.6\\ \midrule
    Data source:    &  & &        \\
    Consultant records         & 25,341,578 & 46.0 & 117.3 \\
    Encyclopedias & 364,066 & 11.5 & 540.4 \\
    Knowledge bases & 798,444 & 15.8 & 35.9 \\ \midrule
    All& 26,504,088 & 44.6 & 120.7\\ \bottomrule
    \end{tabular}
    \caption{Basic statistics  of Huatuo-26M.}
    \label{tab:composition}
\end{table}

\section{Data Statistics and Analysis}

The basic statistics of Huatuo-26M are shown in Table \ref{tab:composition}, most of the QA pairs are from online consultation records. The average length of the dataset questions is 44.6 and the average length of the answers is 120.7. Questions could be long (e.g. in consultant records) or short (in encyclopedias and knowledge bases). There exists both long answers (e.g., Encyclopedia) and short answers (e.g. consultant records and knowledge bases).  We randomly take 1\% QA pairs as the test set while others form the training set.


\begin{figure}[!htb] 
\includegraphics[width=0.44\textwidth]{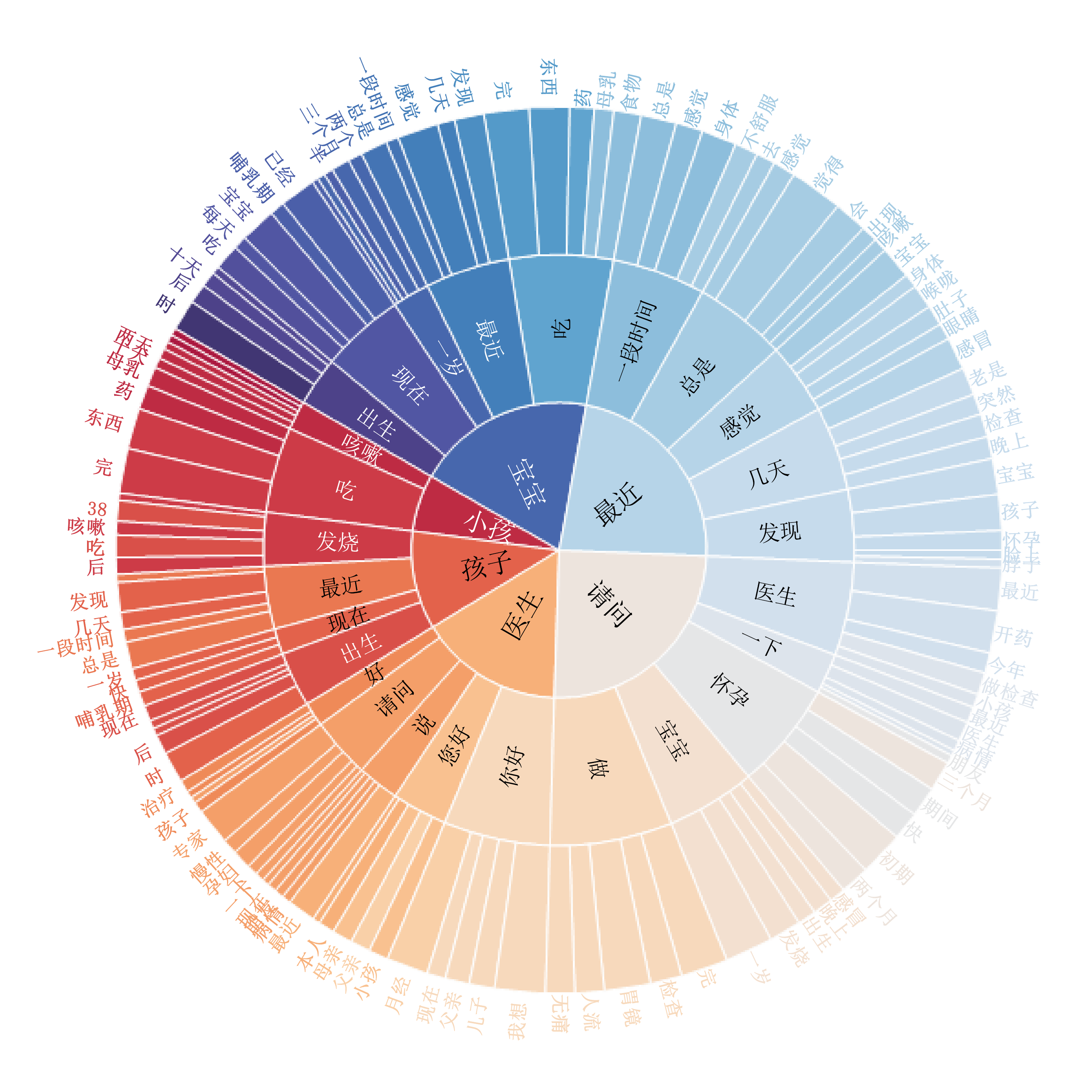}
\centering 
\caption{Distribution of patient-focused questions. We do the analysis heuristically, counting from the first meaningful phrase of the question. We present the relative distribution of these recurring problems and their subsequent distributions.} 
\label{Fig:keyword} 
\end{figure}

\textbf{Questions are colloquial while answers are professional}
 Huatuo-26M consists of a large number of colloquial QA pairs, which are closer to the offline medical diagnosis and contain a lot of medical knowledge. As shown in the sample from online medical consultation in Table \ref{tab:dataset_example}, the patient's question contains patient characteristics and daily symptoms accompanied by life-like scenes, while the doctor's answers are targeted and with contextual semantic continuity.

\textbf{Questions are diverse}
To better understand the characteristics of the data set, we perform heuristic analysis on questions, counting from the first meaningful phrase of the question. As shown in Figure \ref{Fig:keyword}, we get the relative distribution of recurring questions and their subsequent vocabulary distribution. A large part of the patient's question is about newborns, the pregnant, and children, which does make sense given the inexperienced parents and pre-established immunity of children who are prone to illness. At the same time, consultations performed on behalf of elderly parents also accounted for a considerable part, which shows that online consultation has helped solve the problem of China's aging population to a certain extent.

\textbf{Significant topics in Huatuo-26M}
In order to understand the data set in a more fine-grained manner, we also draw word clouds for different data sources in Appendix \ref{appendix:dataset_wordcloud}. We found that for the QA pairs derived from online medical consultation records, the questions cover not only common diseases such as colds and coughs, but also chronic diseases such as hypertension and diabetes, and even a small number of complex diseases such as cysts and epilepsy. The answers focused on prescribing medicines, providing precautions for diet and daily life, and recommending further treatment at the hospital. For the QA pairs from Wikipedia, compared with the QA pairs from consultant records, the proportions of Andrology, Gynecology, and Infectious Diseases have increased. For the QA pairs from the knowledge base, the proportion of questions about intractable diseases such as lymphoma and leukemia increases, while the answers include more professional diagnosis and treatment measures that need to be performed in offline hospitals such as MRI, CT, and puncture.







\begin{table*}[!h]\footnotesize
\centering
\begin{tabular}{llrrrrrrr}
\toprule
Data source                 & Model   & Recall  @5 & Recall  @20 & Recall  @100 & Recall  @1000 & MRR  @10 \\ \midrule
         & BM25    & 4.91     & 6.99      & 10.37       & 17.97        & 3.82   \\
Medical consultant records                           & DeepCT  & \textbf{7.60} & 10.28   & 14.28       & 22.85        & \textbf{6.06}   \\
                           & DPR     & 6.79     & \textbf{11.91} & \textbf{20.96} & \textbf{42.32} & 4.52   \\
                           \midrule
 & BM25    & 4.58     & 8.71      & 17.82       & 39.91        & 3.10   \\
Encyclopedias                           & DeepCT  & \textbf{20.33} & 26.92  & 36.61       & 53.41        & \textbf{16.25}   \\
                           & DPR     & 16.01     & \textbf{27.25} & \textbf{45.33} & \textbf{78.30} & 11.20   \\
                           \midrule
           & BM25    & 0.52     & 1.02      & 1.82       & 3.51        & 0.38   \\
Knowledge bases                            & DeepCT  & 1.05     & 1.46      & 2.10       & 3.29        & 0.71   \\
                           & DPR     & \textbf{2.66} & \textbf{5.25} & \textbf{11.84} & \textbf{33.68} & \textbf{1.83} \\
                           \midrule
                        & BM25    & 4.77     & 6.83      & 10.21       & 17.84        & 3.71   \\
ALL                           & DeepCT  & \textbf{7.58} & 10.24   & 14.22       & 22.68        & \textbf{6.04}   \\
                           & DPR     & 6.79     & \textbf{11.92} & \textbf{21.02} & \textbf{42.55} & 4.53   \\
                           \bottomrule
\end{tabular}
\caption{Retrieval-based benchmark the Huatuo-26M dataset. Results are separated for different data sources. } 
\label{tab:ir_results} 
\end{table*}

\section{Benchmarks}
We benchmarked some mainstream methods on  Huatuo-26M dataset from both answer retrieval and generation.

\subsection{Retrieval based benchmark}
In this section, we will benchmark mainstream retrieval methods on  Huatuo-26M dataset, including both sparse and dense retrieval methods.

\subsubsection{Baselines and Experimental Settings} 
For a given question, we rank the top 1000 relevant answers from the answer pool, which consists of answers from both training and test sets. 
For encyclopedias and knowledge bases, we use 90\% questions for training and the rest for testing. For consultant records or all categories, we use 99\% questions for training and the rest for testing, since testing with 1\% questions is enough and could save more evaluation time than that with  10\% questions.

\paragraph{BM25} BM25 is a bag-of-words retrieval function that ranks a set of documents based on the query terms appearing in each document. Considering that the pre-training models for Chinese are all based on single characters, for the convenience of comparison, we use single characters as units to build indexes instead of words. We utilize the Lucene code base and set k1 to 1.2 and b to 0.9.
\paragraph{DeepCT}  ~\cite{dai2020context} uses BERT \footnote{ We use the BERT checkpoint released by Google, which could be downloaded from https://huggingface.co/bert-base-chinese } to determine context-aware term weights. We trained the model for 3 epochs, with a learning rate of $2\times 10^{-5}$ using Adam. The batch size is set to 72 and the max sequence length is set to 256.
\paragraph{DPR} ~\cite{karpukhin2020dense} learns embeddings by a simple dual encoder framework. The DPR model used in our experiments was trained using the batch-negative setting with a batch size of 192 and additional BM25 negatives. We trained the question and passage encoders for 2 epochs, with a learning rate of $10^{-5}$ using Adam, linear scheduling with warm-up and dropout rate 0.1.

\paragraph{Evaluation Metrics}
We use Recall@k and MRR@10 as evaluation indicators. Recall@k measures the percentage of top k retrieved passages that contain the answer. MRR@10 calculates the average of the inverse of the ranks at which the first relevant document was retrieved.

\subsubsection{Results}

The experimental results are shown in Table \ref{tab:ir_results}. Both DeepCT and DPR outperform BM25, evidencing the effectiveness of neural IR models. In most cases, DPR performs better than DeepCT, this is probably because dense IR models might be generally more powerful than sparse neural IR models. 
Note that the recall performance is relatively low in experiments involving consultant records since the pool of retrieval candidates (i.e., 26M) is too large to recall desired documents. Interestingly, we found that the top-ranked answers are still informative even if it does not recall the desired answer. For specific sample analysis, please refer to App.~\ref{appendix:retrieval_example}.

These retrieval models generally do not perform well in QA extracted from knowledge bases. Since questions in knowledge bases are concise and it requires models to deeply understand knowledge (e.g. medical entities and their in-between relationship). Knowledge representation in pre-trained language models (e.g. in retrieval scenarios) is still challenging; while it becomes more challenging in the medical domain since it is more knowledge-intensive.

{\color{red}
\warningsign} It is worth noting that retrieval-based solutions for medical QA assume that 1) there should be pre-defined answers for all medical questions; and 2) answers should be static for a given question and independent of the different backgrounds of patients. The two assumptions sometimes do not hold. First, there are always some new emergent situations in the medical domain, e.g. COVID-19, which people have little information about it when it just emerges. Second, the answers, e.g., suggestions and treatment, for a given medical question is dependent on the individual’s situation, e.g., age and gender, symptoms and complications, and whether the symptoms are early or late. Therefore, a static answer might not be enough for medical consultation.

\subsection{Generation Based Benchmark}

We fine-tune generative language models (e.g., T5 and GPT2) using the training set of Huatuo-26M and evaluate them in the test set.

\subsubsection{Baselines and Experimental Settings}
\label{sec:generation_setting}

\begin{table*}[!ht]\footnotesize
\centering
\addtolength\tabcolsep{-4.5pt} 
\begin{tabular}{lrrrrrrrrrrr}
\toprule

Model & BLEU-1 & BLEU-2 & BLEU-3 & BLEU-4 & GLEU  & ROUGE-1 & ROUGE-2 & ROUGE-L & Distinct-1 & Distinct-2 \\ \midrule
T5 &	0.33 &	0.18 &	0.12 &	0.07 &	0.10 &	0.67 &	0.19 &	0.63	 &0.01 &	0.02 \\
T5 (fine-tuned)&\textbf{26.63} &\textbf{16.74} &\textbf{11.77} &\textbf{8.46} &	\textbf{11.38} &\textbf{33.21} &\textbf{13.26} &\textbf{24.85} &\textbf{0.51} &\textbf{	0.68} \\
GPT2&	10.04 &	4.60 &	2.67 &	1.62 &	3.34 &	14.26	 &3.42 &	12.07 &	0.17 &	0.22 \\
GPT2 (fine-tuned) &	23.42 &	14.00 &	9.35 &	6.33	 & 9.47 &	30.48 &	11.36 &	23.15 &	0.43 &	0.58 \\
 \bottomrule
\end{tabular}
 \caption{Generation based benchmark on Huatuo-26M.} 
\label{tab:generation_results} 
\end{table*}

We report results for \textit{raw} T5 and GPT2 and the results after \textit{fine-tuning} on  Huatuo-26M train set.
\paragraph{T5}  trains many text-based language tasks in a  unified text-to-text framework. We continuously train T5 for 1 epoch on the full training set using batch-size 8, with a learning rate of $10^{-4}$  using Adam, linear scheduling with a warm-up rate of 0.1. The Chinese T5 model has 12 layers T5~\footnote{\url{https://huggingface.co/imxly/t5-pegasus}}.
\paragraph{GPT2} is a decoder-only generative language model. We fine-tune  GPT2  for 1 epoch on the full training set with a batch size of 12, with a learning rate of $10^{-4}$  using Adam, linear scheduling with a warm-up rate of 0.1. In both T5 and GPT2, the maximum lengths of questions and answers are set to 256 and 512. The Chinese GPT is the original 12-layer GPT2~\footnote{downloaded from \url{https://huggingface.co/uer/gpt2-chinese-cluecorpussmall}}.

\paragraph{Evaluation Metrics}
 We use BLEU, ROUGE, GLEU, and Distinct as evaluation indicators. \textbf{BLEU} evaluates the similarity of generated and reference sentences by computing the k-gram overlap between the generated utterance and the reference. \textbf{ROUGE-N} measures the N-gram overlap between the generated sentence and the reference, and ROUGE-L measures the longest sequence of word matches using the longest common subsequence. \textbf{GLEU} automatically evaluates sentence-level fluency by examining different parsers. \textbf{Distinct-1/2} is an auxiliary metric for evaluating the textual diversity of the generated response by calculating the number of distinct n-grams.

\subsubsection{Results}

The results of the generation benchmark are summarized in Table \ref{tab:generation_results}. Obviously, the fine-tuned T5  and GPT2  models have improved significantly compared to the raw T5 and GPT2 models without fine-tuning, especially fine-tuned  T5  has achieved the best results in all evaluation indicators.
Note the performance of the generation method seems relatively weak (with relatively low scores in these generation metrics), this is probably because the expected answers are typically long and it is more difficult to generate exactly-same long answer than a short answer (like entities in some general QA tasks, e.g. Natural Questions~\cite{kwiatkowski2019natural}). 

{\color{red}
\warningsign} We  warn that generation-based medical QA is risky. Since it is difficult to verify the correctness of generated content; misleading information in the medical domain might lead to more severe ethic issues.  We benchmark these generation methods because generation methods in QA are nowadays more promising than retrieval methods thanks to the success of ChatGPT. However, they are not ready to be deployed in the real world.

\section{Applications}

This section will demonstrate the usefulness of the proposed dataset from many aspects: transfer for other QA datasets, as external knowledge, and as a pre-training corpus in Sec.~
\ref{sec:transfer},~\ref{sec:rag}, and~\ref{sec:bert}.

\begin{table*}[!ht]\footnotesize
\centering
\addtolength\tabcolsep{-4.5pt} 
\resizebox{\linewidth}{!}{
\begin{tabular}{llcccccccccc}
\toprule
Dataset & Model & BLEU-1 & BLEU-2 & BLEU-3 & BLEU-4 & GLEU  & ROUGE-1 & ROUGE-2 & ROUGE-L & Distinct-1 & Distinct-2 \\ \midrule
\multirow{4}{*}{\textbf{cMedQA2}}
&GPT2 (raw)  & 9.96    & 4.30    & 2.33    & 1.33    & 3.18  & 13.85    & 3.07     & 11.60    & 0.175& 0.218\\
&T5 (raw)    & 0.23    & 0.12    & 0.07    & 0.04    & 0.07  & 0.53     & 0.13     & 0.50     & 0.014& 0.015\\
&GPT2 (fine-tuned by Huatuo-26M) & 23.34   & 13.27   & 8.49    & 5.55    & 8.97  & 29.10    & 9.81     & 21.27    & 0.462& 0.611\\ 
&T5 (fine-tuned by Huatuo-26M)  & \textbf{25.65} & \textbf{14.94} & \textbf{9.79} & \textbf{6.64} & \textbf{10.03} & \textbf{30.64} & \textbf{10.49}    & \textbf{21.48} & \textbf{0.543} & \textbf{0.727}\\
\hline
&T5 (fine-tuned by cMedQA2) $^ \dagger$ & 20.88   & 11.87   & 7.69    & 5.09    & 7.62  & 27.16    & 9.30     & 20.11    & 0.418& 0.526\\
\midrule
\multirow{4}{*}{\textbf{webMedQA}} &GPT2 (raw) & 7.84    & 3.51    & 1.99    & 1.16    & 2.56  & 12.00    & 2.70     & 10.07    & 0.120& 0.150\\
&T5 (raw) & 0.47    & 0.21    & 0.13    & 0.08    & 0.13  & 1.04     & 0.20     & 0.97     & 0.009& 0.009\\
&GPT2 (fine-tuned by Huatuo-26M)   & 19.99   & 11.54   & 7.51    & 4.97    & 7.80  & 28.19    & 9.69     & 21.30    & 0.363& 0.494\\
&T5 (fine-tuned by Huatuo-26M) & \textbf{23.20} & \textbf{13.80} & {9.21}& {6.29}  & \textbf{9.22} & {30.68} & \textbf{10.90} & {22.26}   & \textbf{0.462} & \textbf{0.633}\\
\hline
&T5 (fine-tuned by webMedQA)  $^ \dagger$  &  21.42   & 13.79   & \textbf{10.06}   & \textbf{7.38}    & 8.94  & \textbf{31.00 }   & \textbf{13.85}    & \textbf{25.78}    & 0.377& 0.469\\
 \bottomrule
\end{tabular}}
 \caption{Zero-shot performance of models trained on Huatuo-26M. $^ \dagger$ indicates fine-tuning while others are zero-shot.} 
\label{tab:zero_results} 
\end{table*}

\subsection{Transfer for Other QA Dataset}
\label{sec:transfer}
In this section, we will explain how Huatuo-26M is beneficial to the existing QA dataset. 
\paragraph{Problem Setting}
In this section, we directly apply the model pre-trained on the Huatuo-26M dataset and evaluate it on other answer generation datasets. A similar configuration could be found in T5-CBQA ~\cite{2020t5cqba}. 


\paragraph{Experimental Settings} 
We selected two existing Chinese medical QA datasets as examples, namely cMedQA2~\cite{8548603} and webMedQA~\cite{he2019applying}. 
\textbf{cMedQA2} is a publicly available dataset based on Chinese medical questions and answers consisting of 108,000 questions and 203,569 answers. \textbf{webMedQA} is a real-world Chinese medical QA dataset collected from online health consultancy websites consisting of 63,284 questions.
We select the correct QA pairs from these two datasets to train our generation model. 
The model settings of T5 and GPT2 follow Sec.~\ref{sec:generation_setting}.

\paragraph{Results} As shown in Table \ref{tab:zero_results}, the performance of the model pre-trained on the Huatuo-26M dataset is much higher than the raw models.  Especially, additionally training on Huatuo-26M  improves the raw T5 models with 25.42 absolute points in cMedQA2 and   22.73 absolute points in webMedQA.  Moreover, in cMedQA2 dataset, T5 trained in Huatuo-26M which never sees neither the training set nor test of cMedQA2, outperforms T5 trained by cMedQA2  in terms of BLEU-1. This evidences that  Huatuo-26M  includes a wide range of medical knowledge, which is beneficial for downstream medical tasks. Moreover, using Huatuo-26M as a training set achieves better performance on cMedQA2 than using its own training set, this is probably due to the large scale of Huatuo-26M that might have related information in cMedQA2. This shows a great potential of   Huatuo-26M for transfer learning in Chinese medicine.

\subsection{As an External Knowledge}
\label{sec:rag}
\paragraph{Problem Setting} RAG ~\cite{lewis2020retrieval}  combines pre-trained parametric and non-parametric memory (i.e.,  external knowledge) for generation, by doing which memorization can be decoupled from generalization. Here we use the Huatuo-26M as the external knowledge resource in RAG.  For a given question $q$, we use trained DPR as a retrieval model to get the top-ranked QA pair $(q_\textrm{aug},a_\textrm{aug})$ from the QA dataset as an additional input. 

\begin{table*}[!ht]\footnotesize
\centering
\vspace{-5pt}
\addtolength\tabcolsep{-4.5pt} 
\begin{tabular}{lcccccccccc}
\toprule
Model & BLEU-1 & BLEU-2 & BLEU-3 & BLEU-4 & GLEU  & ROUGE-1 & ROUGE-2 & ROUGE-L & Distinct-1 & Distinct-2 \\ \midrule
\multicolumn{3}{l}{\textbf{cMedQA2 Fine-tuned}}\\
T5  & 20.88   & 11.87   & 7.69    & 5.09    & 7.62  & 27.16    & 9.30     & 20.11    & 0.418& 0.526\\
T5-RAG    & 25.86   & 18.48   & 15.26   & 13.02   & 14.27 & 34.24    & 17.69    & 27.54    & 0.395& 0.516\\
T5(Huatuo-26M)& 28.76   & 17.08   & 11.67   & 8.41    & 10.45 & 29.79    & 10.23    & 20.68    & \textbf{0.647}& \textbf{0.831}\\
T5(Huatuo-26M)-RAG & \textbf{31.85} & \textbf{22.77} & \textbf{18.70} & \textbf{15.96} & \textbf{17.08} & \textbf{37.01} & \textbf{19.23} & \textbf{28.72}   & 0.573& 0.760\\
\midrule
\multicolumn{3}{l}{\textbf{webMedQA Fine-tuned}}\\
T5    & 21.42   & 13.79   & 10.06   & 7.38    & 8.94  & 31.00    & 13.85    & 25.78    & 0.377& 0.469\\
T5-RAG& 20.30   & 13.29   & 9.97    & 7.61    & 9.40  & 32.40    & 14.88    & 27.25    & 0.285& 0.377\\
T5(Huatuo-26M)  & \textbf{31.47} & \textbf{20.74} & \textbf{15.35} & \textbf{11.60} & \textbf{12.96} & 34.38    & 15.18    & 26.72    & \textbf{0.651} & \textbf{0.832}\\
T5(Huatuo-26M)-RAG    & 25.56   & 16.81   & 12.54   & 9.58    & 11.80 & \textbf{34.88} & \textbf{15.59}  & \textbf{27.43}    & 0.447& 0.611\\

 \bottomrule
\end{tabular}
 \caption{The comparison  with or without using   Huatuo-26M as an external RAG corpus. The difference with Tab.~\ref{tab:zero_results} is that here we finally fine-tune these models in the target datasets.} 
\label{tab:rag_results} 
\end{table*}

\paragraph{Experimental Setting}


Considering that T5 performs better in zero-shot scenarios than GPT2, we use T5 instead of GPT2 to generate the answer conditioning on a concatenated text $(q_\textrm{aug},a_\textrm{aug},q)$. Since  RAG models rely a retrieval model, we first train a Chinese DPR  model using our dataset. Then we use the document encoder to compute an embedding for each document, and build a single MIPS index using FAISS ~\cite{DBLP:journals/corr/JohnsonDJ17}  for fast retrieval. In RAG training, we retrieve the closest QA pair for each question and split it into $(q_\textrm{aug},a_\textrm{aug},q)$ format. We define the maximum text length after splicing as 400, train for 10 epochs with batch size 24 and learning rate 3e-05.  The difference between \textbf{T5} and \textbf{T5 (Huatuo-26M)} is that the latter was first trained in  Huatuo-26M dataset before training in the target dataset (i.e.,  cMedQA2 or webMedQA).

\paragraph{Results} As shown in Table \ref{tab:rag_results}, we find that the RAG strategy improves the quality of text generation to a certain extent. Particularly, on cMedQA2, the model can consistently benefit from the RAG strategy with and without pre-training on the Huatuo-26M dataset. 
For RAG, we could  additionally  train  backbone models in Huatuo-26M before fine-tuning, as introduced in Sec.~\ref{sec:transfer}; the improvement of the dditional pre-training could be found in cMedQA2 (3 absolute point improvement over purely RAG) but not in webMedQA (nearly 6  absolute point decrease); this might depend on the characteristics of  target datasets.


\subsection{As a Pre-trained Corpus}
\label{sec:bert}

\begin{table*}[!h]
\small
\begin{tabular}{lccccccccc}
\toprule
Model           & CMedEE & CMedIE & CDN  & CTC  & STS  & QIC  & QTR  & QQR  & \textbf{Avg-ALL} \\ \midrule
BERT-base       & \textbf{62.1}   & \textbf{54.0}   & 55.4 & 69.2 & 83.0 & 84.3 & 60.0 & \textbf{84.7} & 69.1     \\
\textbf{BERT-base (Huatuo-26M) }   & 61.8   & 53.7   & \textbf{56.5} & \textbf{69.7} & \textbf{84.6} & \textbf{86.2} & \textbf{62.2} & \textbf{84.7} & \textbf{69.9}     \\
\hline
RoBERTa-base  & 62.4 & 53.7   & 56.4 & 69.4 & 83.7 & 85.5 & 60.3 & 82.7 & 69.3     \\
RoBERTa-large  & 61.8   & \textbf{55.9} & 55.7 & 69.0 & \textbf{85.2} & 85.3 & \textbf{62.8}& 84.4 & 70.0    \\ 
\textbf{RoBERTa-base (Huatuo-26M)} & \textbf{62.8}   & 53.5   & \textbf{57.3} & \textbf{69.8} & 84.9 & \textbf{86.1} & 62.0 & \textbf{84.7} & \textbf{70.1}     \\
\hline
ZEN ~\cite{Diao2019ZENPC} & 61.0 & 50.1 & 57.8 & {68.6} & 83.5 & 83.2 & 60.3 &  83.0 &68.4 \\
MacBERT ~\cite{cui-etal-2020-revisiting}   & 60.7 & 53.2 & 57.7 & 67.7 &   84.4 & 84.9& 59.7 &  {84.0} &69.0 \\
MC-BERT  ~\cite{zhang2020conceptualized} & {61.9} & {54.6} & {57.8} & 68.4 & 83.8 & {85.3} & 61.8 & 83.5 &{69.6} \\ 
\bottomrule
\end{tabular}
\caption{The performance on the test set of  CBLUE evaluation. We use Huatuo-26M as a pre-trained corpus. The results including Zen, MacBERT, and MC-BERT are from the official website. } 
\label{tab:nlu_results} 
\end{table*}

\paragraph{Problem Setting} We use Huatuo-26M as a pre-trained corpus to continue  training existing pre-trained language models like BERT and RoBERTa.

\subsubsection{Experimental Settings} 
\paragraph{BERT} BERT~\cite{devlin2018bert} is a transformer-based language representation model. 
\textbf{BERT-base} is the original 12-layer BERT and the Chinese BERT is downloaded from \url{https://huggingface.co/bert-base-chinese}.
\textbf{BERT-base (Huatuo-26M)} is the model initialized by \textbf{BERT-base} and continuously trained by the Huatuo-26M dataset using masked language model. We trained the model for 10 epochs with a learning rate $5^{-5}$ with batch size 64. Questions and answers are spliced together, and the maximum length is 256. 

\paragraph{RoBERTa} RoBERTa ~\cite{liu2019roberta} is a better-optimized BERT model. The Chinese Roberta is downloaded from \url{https://huggingface.co/hfl/chinese-roberta-wwm-ext}. \textbf{RoBERTa-base} is with 12 layers and \textbf{Roberta-large }is with 24 layers.  \textbf{RoBERTa-base (Huatuo-26M)} is the model initialized by \textbf{RoBERTa-base} and continuously trained by the Huatuo-26M dataset using masked language model.We trained the model for 10 epochs with a learning rate $5^{-5}$ with a batch size 64. Questions and answers are spliced together, and the maximum length is 256. 

\paragraph{ZEN}  ~\cite{Diao2019ZENPC} a BERT-based Chinese text encoder augmented by N-gram representations that take different character combinations into account during training. ZEN thus combines comprehensive information about character sequences and the words or phrases they contain.

\paragraph{MacBERT}  ~\cite{cui-etal-2020-revisiting} reduces the gap between the pre-training and fine-tuning stages by covering words with a similar vocabulary to it, which is effective for downstream tasks. It replaces the original MLM task with the MLM for correction (Mac) task, and mitigates the difference between the pre-training and fine-tuning stages.

\paragraph{MC-BERT} ~\cite{zhang2020conceptualized} study how the pre-trained language model BERT adapts to the Chinese biomedical corpus, and propose a new conceptual representation learning method that a coarse-to-fine cryptographic strategy is proposed to inject entity and linguistic domain knowledge into representation learning.

\subsubsection{Experimental Data}
We evaluated BERT and RoBERTa trained on the Huatuo-26M dataset on the CBLUE~\cite{zhang-etal-2022-cblue}. CBLUE is the first Chinese medical language understanding evaluation benchmark platform, including a collection of natural language understanding tasks such as named entity recognition, information extraction, and single sentence/sentence pair classification.
\subsubsection{Results}
 As shown in Table \ref{tab:nlu_results}, BERT and RoBERTa trained on the Huatuo-26M dataset have significantly improved the performance of CBLUE. The trained 12-layer RoBERTa(Huatuo-26M) model outperforms the 24-layer Roberta model in terms of average scores,  demonstrating that the Huatuo-26M dataset is rich in medical information. The average score of the RoBERTa-base (Huatuo-26M) model is 0.8 percentage points higher than that of the RoBERTa-base model and 0.5 percentage points higher than that of the MC-BERT-base model.




\section{Conclusion}
In this paper, we propose the largest Chinese medical QA dataset to date, consisting of \textbf{26 Million} medical QA pairs, expanding the size of existing datasets by more than 2 orders of magnitude. At the same time, we benchmark many existing works based on the data set and found that these methods still have a lot of room for improvement in medical QA scenarios. We also demonstrate the possible uses of the dataset in practice. The experimental results show that the dataset contains rich medical knowledge that can be very helpful to existing datasets and tasks. We hope that the Huatuo-26M dataset can not only help promote the research of medical QA, but also practically help doctors and patients.

\section*{Limitation}

{\color{red}
\warningsign} The dataset might contain some wrong medical information since its scale is large with 26M QA pairs and manual checking by experts in nearly impossible in the current stage. 
To better maintain the dataset, we aim to build an online website where clinical doctors or experts could modify these QA pairs.  This might be done by recruiting part-time doctors to first check these data and regularly update them. 

This dataset might be translated into other languages, especially low-resource languages. Note that the translation might introduce some additional errors. Moreover, one should also be noticed some basic differences between traditional Chinese medicine and western medicine.

For medical consultation, the treatment/suggestions vary from person to person. In other words, it might be highly dependent on the individual's situation, e.g., age and gender, whether the main symptoms such as pain are accompanied by other symptoms, or whether the symptoms are early or late. The information might need to be confirmed in a multi-turn dialogue instead of single-turn QA. In the future,  we would explore dialogue systems for medical QA. 

\section*{Ethics Statement}

As we mentioned in the limitation, the  collected data might  still have wrong medical information, which comes from two aspects: 1) doctors might make mistakes in online medical consultation, especially given the fact patience might expose incomplete information; and 2) the automatic extraction of QA pairs might also introduce some inaccurate information. Although the data scale is too large to manually check by medical experts, we have made some efforts to reduce its negative effects.  We have highlighted these concerns in many parts of this paper and  warned readers.

\section*{Dataset Download}

All data are crawled from open-source resources. For these data resources where we extract question-answering pairs, namely online encyclopedias, and knowledge bases,  we directly provide full-text question-answering pairs. For the raw data we crawled as question-answering pairs, like online consultation records, we provide two versions: a \textbf{URL version} that provides a URL website associated with a question-answering pair; and a \textbf{full-text version} that directly provides full texts for question-answering pairs. Huatuo-26 providing URL links for online consultation records is fully open-sourced \footnote{The temporary download link is in \url{https://drive.google.com/file/d/1SKsU8owLt3IWZPLlnPytpCwm8-EH3iW6/view},  QA pairs from encyclopedias and knowledge bases are full-text and complete, but one has to crawl QA pairs from online medical consultation records by itself. This is to avoid data misuse from some companies or individuals.}. While Huatuo-26 provides full texts for all QA pairs is only open-sourced to research institutes or universities if they agree on a license to promise for the purpose of research only.

\bibliography{anthology,custom}
\bibliographystyle{acl_natbib}

\appendix



\section{ Word Clouds for Huatuo-26M Dataset}
\label{appendix:dataset_wordcloud}
\begin{figure*}[!htb] 
\centering 
\includegraphics[width=1.0\textwidth]{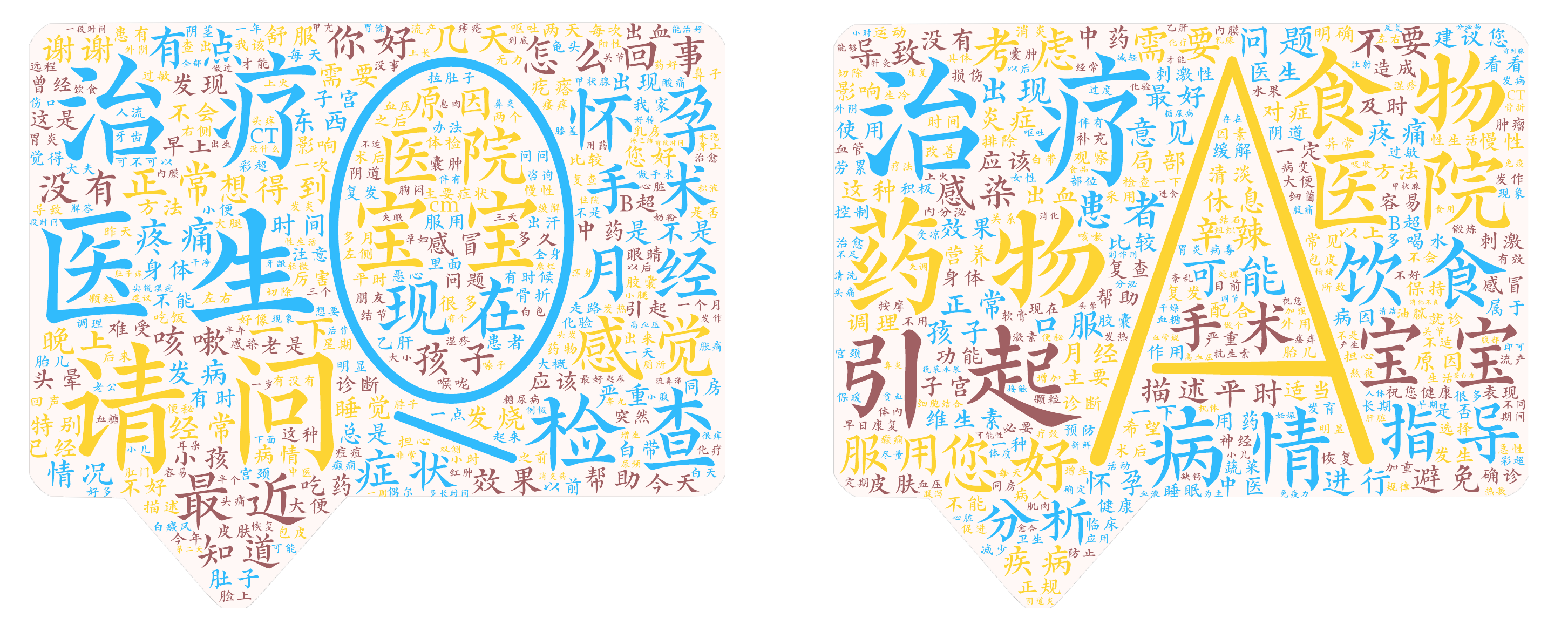}
\caption{Word clouds drawn from Q\&A pairs from online consultation records.}
\label{Fig:consult_wordcloud} 
\end{figure*}

\begin{figure*}[!htb] 
\centering 
\includegraphics[width=1.0\textwidth]{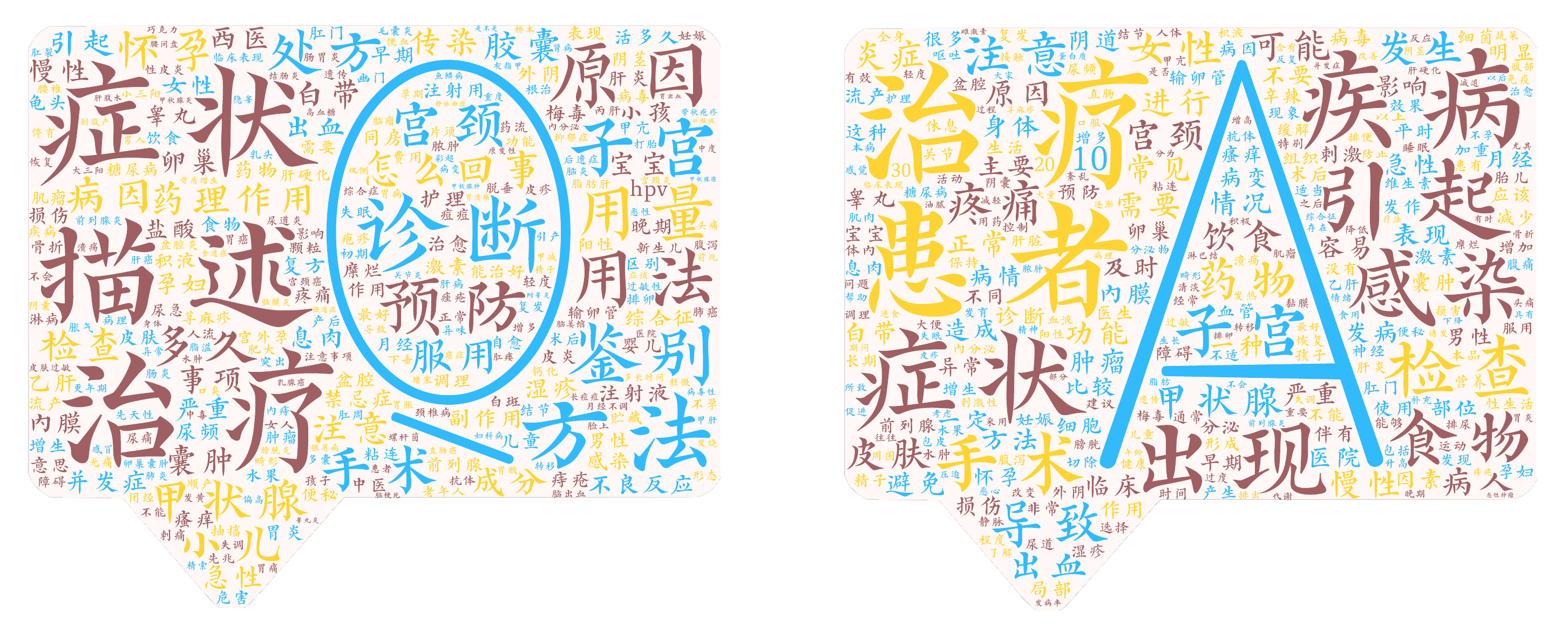}
\caption{Word clouds drawn from Q\&A pairs from Encyclopedia.}
\label{Fig:wiki_wordcloud} 
\end{figure*}

\begin{figure*}[!htb] 
\centering 
\includegraphics[width=1.0\textwidth]{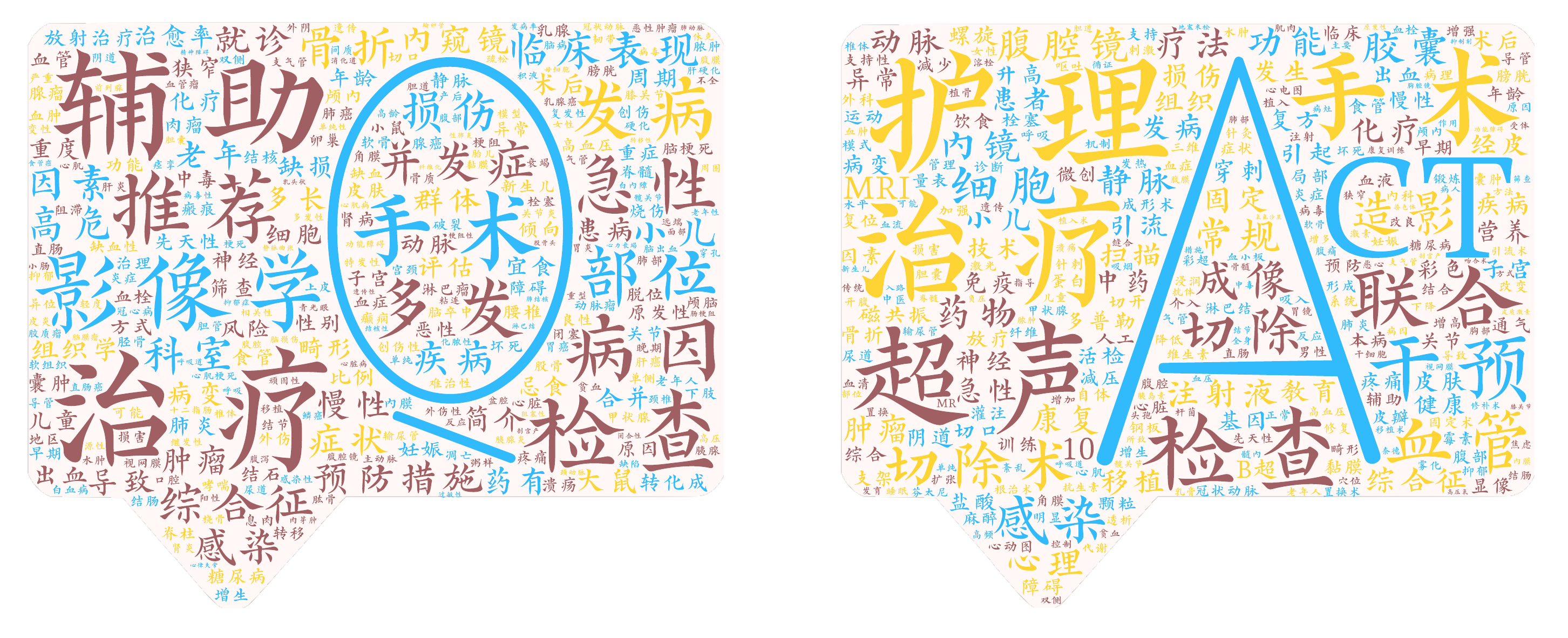}
\caption{Word clouds drawn from Q\&A pairs from Knowledge bases.}
\label{Fig:kb_wordcloud} 
\end{figure*}

As shown in Figure \ref{Fig:consult_wordcloud}, \ref{Fig:wiki_wordcloud}, and \ref{Fig:kb_wordcloud}, we extracted the top 1000 keywords based on TF-IDF and drew word clouds for different sources of Huatuo-26M. It shows QA pairs from online consultation records are more informal since they use more daily words like \begin{CJK}{UTF8}{gbsn}`宝宝' \end{CJK} (namely `a lovely nickname for babies'); while they are more formal in other resources with more professional medical words, the combination between formal and informal questions making this dataset diverse.

\section{Examples of Huatuo-26M Dataset}
\label{appendix:dataset_example}

\begin{table*}[htb]\footnotesize
    \begin{tabular}{cl}
    \toprule
    & \textbf{From Online Medical Consultant records:}  \\
    \midrule
    Question & \begin{CJK}{UTF8}{gbsn}我可能得了戊肝，因为我饮食越来越不好，而且吃饭的时候总是想吐。问一下戊肝早期的症状是什么呢？\end{CJK}\\
             & I may have hepatitis E because my diet is getting worse and I always feel sick when I eat. What are the early \\
             & symptoms of hepatitis E? \\
    Answer   & \begin{CJK}{UTF8}{gbsn}肝早期的症状会出现脂肪无法消化，造成大量脂肪存积于体内，同时胃功能也会紊乱，就会出现厌食，\end{CJK}\\
             & \begin{CJK}{UTF8}{gbsn}看见油腻的食物就恶心的症状，肤色素沉着，指甲颜色改变，毛发改变等。\end{CJK}\\
             & The early symptoms of the liver will be that the fat cannot be digested, causing a large amount of fat to accumulate in\\
             & the body. At the same time, the stomach function will also be disordered, there will be symptoms of anorexia and nau-\\
             & sea when seeing greasy food. Symptoms such as hyperpigmentation and changes in nail color may also occur.\\
    Question & \begin{CJK}{UTF8}{gbsn}3岁宝宝把整个水果糖咽了，怎么才能知道是咽下去了呢？ \end{CJK}  \\
             & The 3-year-old baby swallowed the whole fruit candy, how can I know that the baby has swallowed it and \\
             & not stuck it in the throat?  \\
    Answer   & \begin{CJK}{UTF8}{gbsn}只要是咽后宝宝没有憋气的现象，那就是咽下去了。\end{CJK}  \\
             & As long as the baby does not hold his breath after swallowing, the baby has swallowed the thing.\\
             \midrule
    & \textbf{From Online Medical Encyclopedia:}  \\
    \midrule
    Question & \begin{CJK}{UTF8}{gbsn}前列腺钙化灶是怎么治呢？\end{CJK}  \\
             & How is prostate calcification treated? \\
    Answer   & \begin{CJK}{UTF8}{gbsn}钙化灶是X线检测到的前列腺内的钙质沉积。前列腺内的钙化灶有大小之分：粗大的钙化灶常常为列腺内的\end{CJK}  \\
             & \begin{CJK}{UTF8}{gbsn}良性病变，如前列腺内动脉的老化、陈旧性的损伤以及炎症等，不需要进一步活检。细小的钙化灶通常位于\end{CJK}  \\
             & \begin{CJK}{UTF8}{gbsn}细胞生长分裂较快的部分。出现前列腺钙化或结石必须治疗，钙化会发展成结石，引发出各种症状，有的症\end{CJK}  \\
             & \begin{CJK}{UTF8}{gbsn}状长期消除不了，要做全面检查，看是否有结石钙化，不治疗结石钙化难以彻底治愈前列腺病。保养治疗需\end{CJK}  \\
             & \begin{CJK}{UTF8}{gbsn}要劳逸结合，防止过度疲劳进行适当的体育运动，尤其是加强盆腔肌肉的运动，忌长久静坐，忌长久骑车，\end{CJK}  \\
             & \begin{CJK}{UTF8}{gbsn}忌久蹲，排便时间控制在3到5分钟，忌坐潮湿之地。便后清洁肛门。注意饮食，多饮水，忌酒及辛辣食物。\end{CJK}  \\
             & \begin{CJK}{UTF8}{gbsn}多食蔬菜、水果及坚果类食物。因坚果类食物中富含铜和锌，对前列腺有益。\end{CJK}  \\
             & Calcifications are calcium deposits in the prostate that are detected on x-rays. The calcifications in the prostate can be\\
             & divided into different sizes: Coarse calcifications are often benign lesions in the prostate, such as aging of the internal-\\
             & prostatic artery, old injury, and inflammation, and no further biopsy are required. Fine calcifications are usually located\\
             & in the part where the cells are growing and dividing more rapidly. Prostate calcification or stones must be treated. Cal-\\
             & cification will develop into stones and cause various symptoms. Some symptoms cannot be eliminated for a long time. \\
             & A comprehensive examination should be done to see if there are stone calcifications. Prostate disease cannot be comp-\\
             & letely cured without treatment for calcification. Maintenance treatment requires a combination of work and rest to pre-\\
             & vent excessive fatigue and carry out appropriate physical exercises, especially exercises to strengthen pelvic muscles.\\
             & Avoid sitting for a long time, riding a bicycle for a long time, and squatting for a long time. The defecation time is co-\\
             & ntrolled within 3 to 5 minutes. Avoid sitting in wet places. Clean the anus after defecation. Pay attention to diet, drink\\
             & plenty of water, avoid alcohol and spicy food. Eat more vegetables, fruits and nuts. Nuts are rich in copper and zinc, it\\
             & is good for the prostate.\\
    Question & \begin{CJK}{UTF8}{gbsn}什么是生物药剂学？ \end{CJK}  \\
             & The 3-year-old baby swallowed the whole fruit candy, how can I know that the baby has swallowed it and \\
             & not stuck it in the throat?  \\
    Answer   & \begin{CJK}{UTF8}{gbsn}生物药剂学是研究给药后药物的吸收的整个体内过程，包含各种制剂因素和生物因素对这一过程与药效的影\end{CJK}  \\
             & \begin{CJK}{UTF8}{gbsn}响。此外，生物药剂学通过药物对生物细胞产生的反应过程来达到施药者想要达到的目的。1950年代初，人\end{CJK}  \\
             & \begin{CJK}{UTF8}{gbsn}们普遍认为“化学结构决定药效”，药剂学只是为改善外观、掩盖不良嗅味而便于服用。随着大量的临床实\end{CJK}  \\
             & \begin{CJK}{UTF8}{gbsn}践证明，人们逐渐开始认识到剂型和生物因素对药效的影响。因此研究药物在代谢过程的各种机理和理论及\end{CJK}  \\
             & \begin{CJK}{UTF8}{gbsn}各种剂型和生物因素对药效的影响，对控制药物之际的内在品质，确保最终药品的安全有效，提供新药开发\end{CJK}  \\
             & \begin{CJK}{UTF8}{gbsn}和用药的严格评价，都具有重要的意义。\end{CJK}  \\
             & Biopharmaceutics is the study of the entire process of drug absorption after administration, including the effects of var-\\
             & ious preparation factors and biological factors on this process and drug efficacy. Biopharmaceutics uses the process of\\
             & drug response to biological cells to achieve the expected purpose. In the early 1950s, it was generally believed that\\
             & "the chemical structure determines the efficacy of the drug", and pharmacy was only for improving the appearance and\\
             & masking the bad smell to make it easier to take. With a large number of clinical practices, people gradually began to re-\\
             & alize the influence of dosage forms and biological factors on drug efficacy. It's important to study various mechanisms\\
             & and theories of drugs in the metabolic process and the influence of various dosage forms and biological factors on drug\\
             & efficacy, control the internal quality of drugs, ensure the safety and effectiveness of final drugs, and provide strict eval-\\
             & uation for new drug development. \\
             \midrule
    & \textbf{From Online Medical Knowledge bases:}  \\
    \midrule
    Question & \begin{CJK}{UTF8}{gbsn}脓腔穿刺的辅助治疗有些什么？\end{CJK}\\
             & What are the adjuvant treatments for abscess puncture? \\
    Answer   & \begin{CJK}{UTF8}{gbsn}消毒隔离；皮肤的护理；营养支持\end{CJK}\\
             & Disinfection and isolation; skin care; nutritional support\\
    Question & \begin{CJK}{UTF8}{gbsn}气道吸痰的辅助治疗有些什么？ \end{CJK}  \\
             & What are the adjunctive treatments for airway suctioning? \\
    Answer   & \begin{CJK}{UTF8}{gbsn}足量补液\end{CJK}  \\
             & Adequate rehydration\\
             \bottomrule
    \end{tabular}
    \caption{Examples from various sources of the dataset}
    \label{tab:dataset_example}
\end{table*}

Table \ref{tab:dataset_example} shows examples from various sources of the dataset, and the data characteristics of each data source can be roughly seen through the examples. For Q\&A pairs derived from online medical consultation records, the questions are more colloquial and the answers are more targeted. For Q\&A pairs sourced from online medical wikis and expert articles, the questions are more concise, rarely involving specific patient information, and the answers are more detailed and professional. For Q\&A pairs from online medical knowledge bases, the questions are concise, the answers are accurate, and there are fewer identical texts between answers and questions.

\section{Extracting QA pairs from encyclopedia page}
\label{appendix:wiki_bases_templates}

\begin{figure*}[htbp] 
\centering
\includegraphics[width=0.7\textwidth]{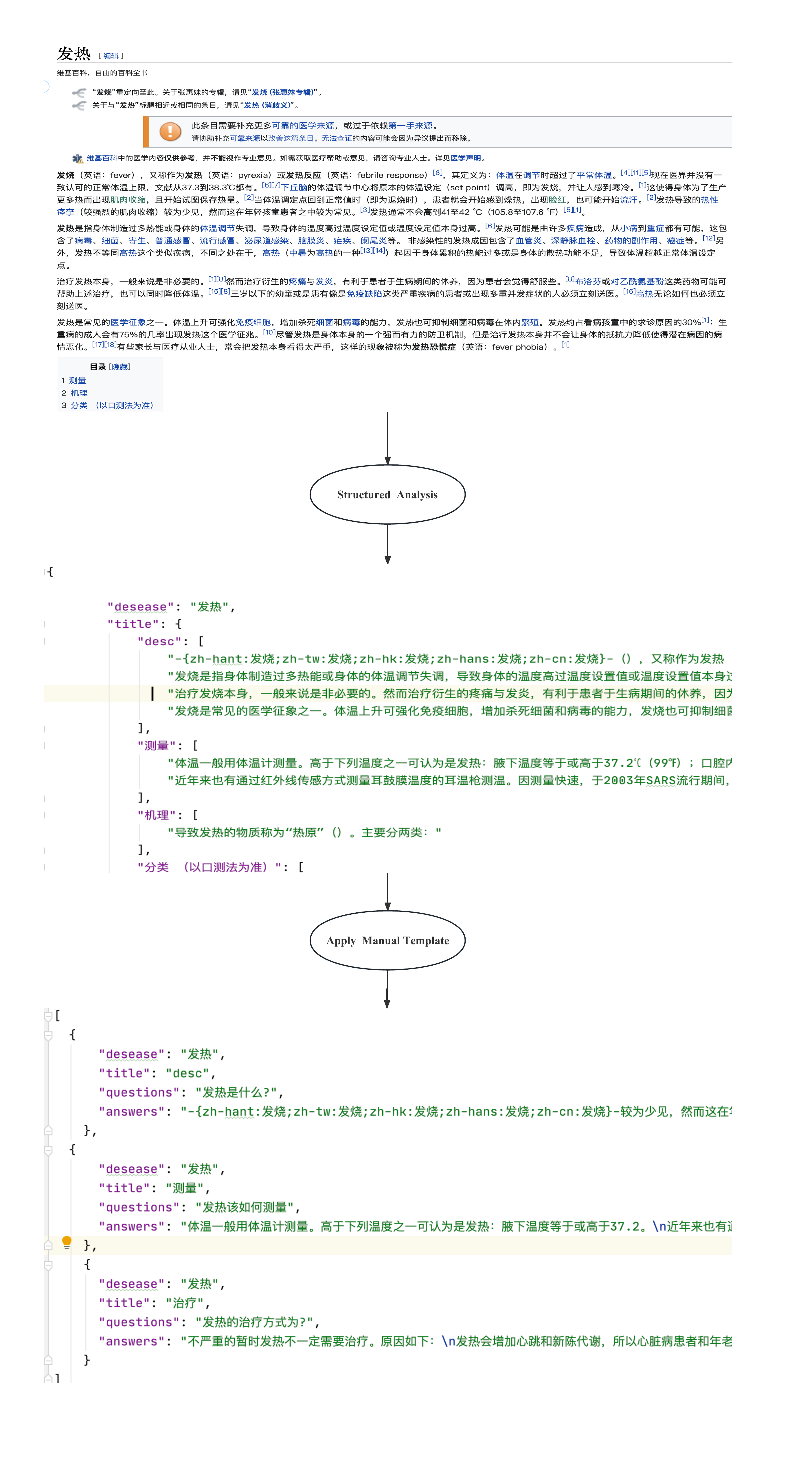}
\caption{Workflow for extracting QA pairs from WIKI pages.} 
\label{Fig:wiki-extract-QA} 
\end{figure*}

As shown Fig.~\ref{Fig:wiki-extract-QA} , For a given Wikipedia page, we use an HTML parsing tool to extract its structured information based on the contents of the page. Therefore, we get a title based on the contents which are associated with one or many paragraphs. 
Next, we transform each title to a question that could be answered by its associated paragraphs, according to a manually-designed template like Tab.~\ref{tab:knowledge_bases_templates}.


\section{Knowledge Bases Questions Templates }
\label{appendix:knowledge_bases_templates}
Tab.~\ref{tab:knowledge_bases_templates}  shows the generated templates for all knowledge graph questions. Each question template is associated with either a relation between entities or an attribute of an entity. Each question template is conditioned on the subject entity, see the placeholder of entities like {\tt disease} and {\tt drug} in Tab.~\ref{tab:knowledge_bases_templates}. Note that the answer to the question should be the object entity or the attribute of the subject entity. There are 43 question types in total.  We manually checked 500 random examples where the `answer' could match the question; the results show nearly every QA pair is correct.


\begin{CJK}{UTF8}{gbsn}
\begin{table*}[!ht]\tiny
\centering
\addtolength\tabcolsep{-1pt}
\begin{tabular}{llllll}
\toprule
疾病（disease） &		症状	 		(symptom)									& {\tt [disease]}的症状是什么？	 		（What are the symptoms of {\tt [disease]}?）				          \\
疾病（disease） &		并发症	 	(complication)								& {\tt [disease]}的并发症是什么？	 	（What are the complications  of {\tt [disease]}?）			          \\
疾病（disease） &		简介	 		(Introduction)								& {\tt [disease]}的简介是？	 		（What is the profile  of {\tt [disease]}?）						          \\
疾病（disease） &		预防	 		(prevention)								& {\tt [disease]}的预防措施有哪些？	 	（What are the preventive measures  of {\tt [disease]}?）		          \\
疾病（disease） &		病因	 		(Etiology)									& {\tt [disease]}的发病原因？	 		（What is the cause of  {\tt [disease]}?）				          \\
疾病（disease） &		发病率	 	(Morbidity)									& {\tt [disease]}的患病比例是多少？	 	（What is the prevalence rate of  {\tt [disease]}?）			          \\
疾病（disease） &		就诊科室	 	(Medical department)						& {\tt [disease]}的就诊科室是什么？	 	（What is the clinic  of {\tt [disease]}?）						          \\
疾病（disease） &		治疗方式	 	(treatment)									& {\tt [disease]}的治疗方式是什么？	 	（What is the treatment  of {\tt [disease]}?）					          \\
疾病（disease） &		治疗周期	 	(treatment cycle)							& {\tt [disease]}的治疗周期多长？	 	（How long is the treatment cycle  of {\tt [disease]}?）			          \\
疾病（disease） &		治愈率	 	(cure rate)									& {\tt [disease]}的治愈率是多少？	 	（What is the cure rate in  of {\tt [disease]}?）				          \\
疾病（disease） &		检查	 		(an examination	)							& {\tt [disease]}的检查有些什么？	 	（Which check are there for {\tt [disease]}?）		          \\
疾病（disease） &		多发群体	 	(Frequent group)							& {\tt [disease]}的多发群体是？	 		（Which group of people is more likely to get {\tt [disease]}?）	          \\
疾病（disease） &		药物治疗	 	(medical treatement	)						& {\tt [disease]}的推荐药有哪些？	 	（What are the recommended drugs for  {\tt [disease]}?）		          \\
疾病（disease） &		忌食	 		(Do not eat)								& {\tt [disease]}忌食什么？	 		（What shouldn't one eat for {\tt [disease]}?）          						\\
疾病（disease） &		宜食	 		(Edible)									& {\tt [disease]}宜食什么？	 		（What should one eat for {\tt [disease]}?）	          					\\
疾病（disease） &		死亡率	 	(death rate)								& {\tt [disease]}的死亡率是多少？	 	（What is the death rate for  {\tt [disease]} ?）		          		\\
疾病（disease） &		辅助检查	 	(Auxiliary inspection)						& {\tt [disease]}的辅助检查有些什么？	（What are the auxiliary inspections  of {\tt [disease]}?）			    \\
疾病（disease） &		多发季节	 	(High season)								& {\tt [disease]}的多发季节是什么时候？	（Which season do people most likely get {\tt [disease]}?）				          \\
疾病（disease） &		相关（症状）	(related (symptoms))						& {\tt [disease]}的相关症状有些什么？ 	（What are the side symptoms of  {\tt [disease]}?）		          \\
疾病（disease） &		发病机制	 	(pathogenesis)								& {\tt [disease]}的发病机制是什么？	 	（What is the pathogenesis of {\tt [disease]}）		         		 \\
疾病（disease） &		手术治疗	 	(operation treatment)						& {\tt [disease]}的手术治疗有些什么？	（What is the surgical treatment of {\tt [disease]}?）			          \\
疾病（disease） &		转移部位	 	(metastatic site)							& {\tt [disease]}的转移部位是什么？	 	（What is the site of transfer for {\tt [disease]}?）		          \\
疾病（disease） &		风险评估因素	(risk assessment factors)					& {\tt [disease]}的风险评估因素有些什么	（What are the risk assessment factors for {\tt [disease]}）？	          \\
疾病（disease） &		筛查	 		(screening)									& {\tt [disease]}的筛查有些什么？	 	（What are the screenings for {\tt [disease]}?）		          		\\
疾病（disease） &		传播途径	 	(way for spreading)							& {\tt [disease]}的传播途径有些什么？	（What are the channels of transmission of {\tt [disease]}?）		          \\
疾病（disease） &		发病部位	 	(Diseased site)								& {\tt [disease]}的发病部位是什么？	 	（What is the site of {\tt [disease]}?）		         		 \\
疾病（disease） &		高危因素	 	(high risk factors)							& {\tt [disease]}的高危因素有些什么？	（What are the high-risk factors for {\tt [disease]}?）			          \\
疾病（disease） &		发病年龄	 	(Age of onset)								& {\tt [disease]}的发病年龄是多少？	 	（What is the age of onset for {\tt [disease]}?）		        		  \\
疾病（disease） &		预后生存率	(prognostic survival rate)					& {\tt [disease]}的预后生存率是多少？	（What is the prognosis for survival for {\tt [disease]}?）		          \\
疾病（disease） &		组织学检查	(Histological examination)					& {\tt [disease]}的组织学检查有些什么？	（What are the histological examinations for {\tt [disease]}?）	          \\
疾病（disease） &		辅助治疗	 	(adjuvant therapy)							& {\tt [disease]}的辅助治疗有些什么？	（What are  adjuvant treatments of {\tt [disease]}?）			          \\
疾病（disease） &		多发地区	 	(High-risk areas)							& {\tt [disease]}的多发地区是哪里？	 	（Where are the frequent occurrence areas of {\tt [disease]}?）	          \\
疾病（disease） &		遗传因素	 	(genetic factors)							& {\tt [disease]}的遗传因素是什么？	 	（What is the genetic factor of {\tt [disease]}?）		         			 \\
疾病（disease） &		发病性别倾向	(Onset sex tendency)						 & {\tt [disease]}的发病性别倾向是啥？	（What is the sex tendency of onset of {\tt [disease]}?）			          \\
疾病（disease） &		放射治疗	 	(Radiation Therapy)							& {\tt [disease]}的放射治疗有些什么？	（What is radiation therapy of {\tt [disease]}?）			         		 \\
疾病（disease） &		化疗	 		(chemotherapy)								& {\tt [disease]}的化疗有些什么？	 	（What is the chemotherapy of {\tt [disease]}?）		          			\\
疾病（disease） &		临床表现	 	(clinical manifestations)					& {\tt [disease]}的临床表现有些什么？	（What are the clinical manifestations of {\tt [disease]}?）		          \\
疾病（disease） &		内窥镜检查	(endoscopy)						 			& {\tt [disease]}的内窥镜检查有些什么？	（What is the endoscopy examination of {\tt [disease]}?）			          \\
疾病（disease） &		影像学检查	(Film degree exam)						 	& {\tt [disease]}的影像学检查有些什么？	（What are  the imaging tests of {\tt [disease]}?）				          \\
疾病（disease） &		相关（导致）	(related (resulting in))					& {\tt [disease]}会导致什么样的结果？	（What consequence does {\tt [disease]} lead to?）		          				\\
疾病（disease） &		治疗后症状	(Symptoms after treatment)					& {\tt [disease]}的治疗后症状是什么？	（What are the symptoms after treatment for {\tt [disease]}?）			          \\
疾病（disease） &		相关（转化）	(relevant (conversion))						& {\tt [disease]}会转化成什么？	 		（What will {\tt [disease]} translate into?）	          						\\
药品（drug） &		推荐药（diseases cured with this drug）	 				& {\tt [drug]}能治理什么疾病?	 	（What diseases can {\tt [drug]} treat?）					          \\
 \bottomrule
\end{tabular}
 \caption{ Templates to transform relations in knowledge bases to questions.} 
\label{tab:knowledge_bases_templates} 
\end{table*}
\end{CJK}

\section{Examples of Retrieval Based Benchmark}
\label{appendix:retrieval_example}
\begin{table*}[htb]\footnotesize
    \centering
    \begin{tabular}{cl}
    \toprule
    Question & \begin{CJK}{UTF8}{gbsn}{\color{blue}前列腺钙化灶是怎么治呢？}\end{CJK}  \\
    Answer   & \begin{CJK}{UTF8}{gbsn}钙化灶是X线检测到的前列腺内的钙质沉积。前列腺内的钙化灶有大小之分：粗大的钙化灶常常为列腺内的\end{CJK}  \\
             & \begin{CJK}{UTF8}{gbsn}良性病变，如前列腺内动脉的老化、陈旧性的损伤以及炎症等，不需要进一步活检。细小的钙化灶通常位于\end{CJK}  \\
             & \begin{CJK}{UTF8}{gbsn}细胞生长分裂较快的部分。出现前列腺钙化或结石必须治疗，钙化会发展成结石，引发出各种症状，有的症\end{CJK}  \\
             & \begin{CJK}{UTF8}{gbsn}状长期消除不了，要做全面检查，看是否有结石钙化，不治疗结石钙化难以彻底治愈前列腺病。保养治疗需\end{CJK}  \\
             & \begin{CJK}{UTF8}{gbsn}要劳逸结合，防止过度疲劳进行适当的体育运动，尤其是加强盆腔肌肉的运动，忌长久静坐，忌长久骑车，\end{CJK}  \\
             & \begin{CJK}{UTF8}{gbsn}忌久蹲，排便时间控制在3到5分钟，忌坐潮湿之地。便后清洁肛门。注意饮食，多饮水，忌酒及辛辣食物。\end{CJK}  \\
             & \begin{CJK}{UTF8}{gbsn}多食蔬菜、水果及坚果类食物。因坚果类食物中富含铜和锌，对前列腺有益。\end{CJK}  \\
    Top 1    & \begin{CJK}{UTF8}{gbsn}如果只是单纯的钙化灶那么不必担心多数可能是以前有过炎症治愈后留下的证据可以先查一个前列腺液常规\end{CJK}  \\
             & \begin{CJK}{UTF8}{gbsn}{\color{blue} 原问题:再去检查一次前列腺实质内钙化灶什么问题？有什么要注意？想治疗好}\end{CJK}\\
    Top 2    & \begin{CJK}{UTF8}{gbsn}你好根据你的情况看来单纯的钙化灶一般不会有什么症状也无需治疗如果合并有前列腺炎需要针对前列腺炎\end{CJK}  \\
             & \begin{CJK}{UTF8}{gbsn}进行治疗可以通过一些抗菌素中药等等进行治疗以及温水坐浴勿劳累过度熬夜避免饮酒辛辣刺激性食物等等\end{CJK}  \\
             & \begin{CJK}{UTF8}{gbsn}同时注意自己的饮食还有休息情况如果严重请到医院就医祝你早日康复\end{CJK}  \\
             & \begin{CJK}{UTF8}{gbsn}{\color{blue}原问题:最近到医院检查身体发现自己的前列腺增生钙化，在很小的时候曾经换过前列腺炎不过后来发现的}\end{CJK}\\
             & \begin{CJK}{UTF8}{gbsn}{\color{blue}及时已经治愈了，现在看来是不是以前留下来的后遗症。请问我这种前列腺增生钙化怎么治疗？可以治愈吗？}\end{CJK}\\
    Top 3    & \begin{CJK}{UTF8}{gbsn}前列腺钙化灶这种情况下是因为有之前炎症导致的疤痕情况，现在要注意再看一下是否还有尿路感染等导致\end{CJK}  \\
             & \begin{CJK}{UTF8}{gbsn}的情况，可以通过药物来进行治疗的。要注意多喝水，不要憋尿，而且要注意避免辛辣刺激性的食物，平时\end{CJK}  \\
             & \begin{CJK}{UTF8}{gbsn}的话注意增强个人体质来改善的。\end{CJK}  \\
             & \begin{CJK}{UTF8}{gbsn}{\color{blue}原问题:18岁前列腺钙化灶怎么办？}\end{CJK}\\
             \midrule
    Question & \begin{CJK}{UTF8}{gbsn}{\color{blue}什么是生物药剂学？} \end{CJK} \\
    Answer   & \begin{CJK}{UTF8}{gbsn}生物药剂学是研究给药后药物的吸收的整个体内过程，包含各种制剂因素和生物因素对这一过程与药效的影\end{CJK}  \\
             & \begin{CJK}{UTF8}{gbsn}响。此外，生物药剂学通过药物对生物细胞产生的反应过程来达到施药者想要达到的目的。1950年代初，人\end{CJK} \\
             & \begin{CJK}{UTF8}{gbsn}们普遍认为“化学结构决定药效”，药剂学只是为改善外观、掩盖不良嗅味而便于服用。随着大量的临床实\end{CJK}  \\
             & \begin{CJK}{UTF8}{gbsn}践证明，人们逐渐开始认识到剂型和生物因素对药效的影响。因此研究药物在代谢过程的各种机理和理论及\end{CJK}  \\
             & \begin{CJK}{UTF8}{gbsn}各种剂型和生物因素对药效的影响，对控制药物之际的内在品质，确保最终药品的安全有效，提供新药开发\end{CJK}  \\
             & \begin{CJK}{UTF8}{gbsn}和用药的严格评价，都具有重要的意义。\end{CJK}  \\
    Top 1    & \begin{CJK}{UTF8}{gbsn}生物药理学，在生物制药和医药生物技术是跨学科领域之间的药理学和生物技术，认为是一种新兴的科学。\end{CJK}  \\
             & \begin{CJK}{UTF8}{gbsn}包括获得药物的生物起源在生物反应器。一个主要的优势使用这条路线，而不是获得化学合成，避免了消旋\end{CJK}  \\
             & \begin{CJK}{UTF8}{gbsn}的产品，这样就可以获得大量易纯化产品的同类产品，提高性能并降低成本。另一个优势是获得化合物，几\end{CJK}  \\
             & \begin{CJK}{UTF8}{gbsn}乎无法获得任何其他方式尽可能多的重组蛋白。这有助于科学的设计和开发新疗法。\end{CJK}  \\
             & \begin{CJK}{UTF8}{gbsn}{\color{blue}原问题:什么是生物药理学？}\end{CJK}\\
    Top 2    & \begin{CJK}{UTF8}{gbsn}生物产药，又译基因产药术或药耕，是遗传工程学的一种透过基因改造的动植物来生产药物的方法。以此方\end{CJK}  \\
             & \begin{CJK}{UTF8}{gbsn}式生产的通常是重组蛋白质或者其代谢产物。重组蛋白质通常用在生物反应器皿中通过细菌和酵母生产，但\end{CJK}  \\
             & \begin{CJK}{UTF8}{gbsn}是通过生物产药不需要高昂的基础设备，产能可以更加低廉的费用按需而变。\end{CJK}  \\
             & \begin{CJK}{UTF8}{gbsn}{\color{blue}原问题:什么是生物产药？}\end{CJK}\\
    Top 3    & \begin{CJK}{UTF8}{gbsn}原答案\end{CJK}  \\
    \bottomrule
    \end{tabular}
    \caption{Examples of retrieval results of DPR model on questions generated from Encyclopedia}
    \label{tab:retrieval_example_wiki}
\end{table*}

\begin{table*}[htb]\footnotesize
    \centering
    \begin{tabular}{cl}
    \toprule
    Question & \begin{CJK}{UTF8}{gbsn}{\color{blue}脓腔穿刺的辅助治疗有些什么？}\end{CJK}\\
    Answer   & \begin{CJK}{UTF8}{gbsn}消毒隔离；皮肤的护理；营养支持\end{CJK}\\
    Top 1    & \begin{CJK}{UTF8}{gbsn}不留死腔；引流通畅；支管开窗引流；了解脓腔范围\end{CJK}\\
             & \begin{CJK}{UTF8}{gbsn}{\color{blue}原问题:粘膜下脓肿的辅助治疗有些什么？}\end{CJK}\\
    Top 2    & \begin{CJK}{UTF8}{gbsn}保持引流通畅；护理干预；严格拔管；严格无菌操作；保持引流瓶的合适高度\end{CJK}\\
             & \begin{CJK}{UTF8}{gbsn}{\color{blue}原问题:双侧脑室外引流的辅助治疗有些什么？}\end{CJK}\\
    Top 3    & \begin{CJK}{UTF8}{gbsn}破膜；程控降温法；换液；选择合适的供者和采集方法；康复训练；洗血；巴氏消毒；贴壁法；55℃水浴加热\end{CJK}\\
             & \begin{CJK}{UTF8}{gbsn}{\color{blue}原问题:脐血的辅助治疗有些什么？}\end{CJK}\\
    \midrule
    Question & \begin{CJK}{UTF8}{gbsn}{\color{blue}气道吸痰的辅助治疗有些什么？} \end{CJK}  \\
    Answer   & \begin{CJK}{UTF8}{gbsn}足量补液\end{CJK}  \\
    Top 1    & \begin{CJK}{UTF8}{gbsn}加温湿化器联合密闭式吸痰法\end{CJK}\\
             & \begin{CJK}{UTF8}{gbsn}{\color{blue}原问题:NSICU人工气道的辅助治疗有些什么？}\end{CJK}\\
    Top 2    & \begin{CJK}{UTF8}{gbsn}平卧位；床头抬高30°到45°体位；按需吸痰原则\end{CJK}\\
             & \begin{CJK}{UTF8}{gbsn}{\color{blue}原问题:胃肠反流的辅助治疗有些什么？}\end{CJK}\\
    Top 3    & \begin{CJK}{UTF8}{gbsn}常规雾化排痰法；气管按压法；吸痰管刺激法\end{CJK}\\
             & \begin{CJK}{UTF8}{gbsn}{\color{blue}原问题:诱导性主动咳嗽的辅助治疗有些什么？}\end{CJK}\\
             \bottomrule
    \end{tabular}
    \caption{Examples of retrieval results of DPR model on questions generated from Knowledge bases}
    \label{tab:retrieval_example_KB}
\end{table*}

\begin{table*}[htbp]\footnotesize
    \centering
    \begin{tabular}{cl}
    \toprule
    Question & \begin{CJK}{UTF8}{gbsn}{\color{blue}我可能得了戊肝，因为我饮食越来越不好，而且吃饭的时候总是想吐。问一下戊肝早期的症状是什么呢？}\end{CJK}\\
    Answer   & \begin{CJK}{UTF8}{gbsn}肝早期的症状会出现脂肪无法消化，造成大量脂肪存积于体内，同时胃功能也会紊乱，就会出现厌食，\end{CJK}\\
             & \begin{CJK}{UTF8}{gbsn}看见油腻的食物就恶心的症状，肤色素沉着，指甲颜色改变，毛发改变等。\end{CJK}\\
    Top 1    & \begin{CJK}{UTF8}{gbsn}戊型肝炎通常发病比较急，并且在发病期初可能会有恶心，呕吐以及稍稍有一些黄疸的症状。这个疾病主要\end{CJK}\\
             & \begin{CJK}{UTF8}{gbsn}是通过粪口途径传播的，并且常常在老人以及孕妇或者是有乙肝基础的病人发病率比较高。这个疾病通常早\end{CJK}\\
             & \begin{CJK}{UTF8}{gbsn}期应该严格卧床休息，直到症状消失，才可以逐渐正常活动。\end{CJK}\\
             & \begin{CJK}{UTF8}{gbsn}{\color{blue}原问题:我最近听说我朋友得了戊肝，我不太了解这个疾病，这个是不是病毒性肝炎？}\end{CJK}\\
    Top 2    & \begin{CJK}{UTF8}{gbsn}戊型肝炎主要经粪—口途径传播，大多数报道的暴发性流行都与饮用了被粪便污染的水有关，大暴发常常是\end{CJK}\\
             & \begin{CJK}{UTF8}{gbsn}在暴雨与洪水发生之后，水源被污染时出现，多见于秋冬季。也可散发，在环境与水源卫生状况差的地区，\end{CJK}\\
             & \begin{CJK}{UTF8}{gbsn}全年都有散发病例。此外，还可通过日常生活接触和输入性传播。症状可能会出现发热、头痛、咽痛、鼻塞、\end{CJK}\\
             & \begin{CJK}{UTF8}{gbsn}呕吐、上腹不适、肝区痛、腹胀、腹泻等。每个人体质和病情不同，症状就不同。\end{CJK}\\
             & \begin{CJK}{UTF8}{gbsn}{\color{blue}原问题:我最近听说很多人得了戊型肝炎，我也想预防一下，想知道一下戊肝的症状原因？}\end{CJK}\\
    Top 3    & \begin{CJK}{UTF8}{gbsn}戊型其实是由是由肝炎病毒所致的全身性传染病，主要累及肝脏。其临床表现为食欲减退、恶心、乏力、上 \end{CJK}  \\
             & \begin{CJK}{UTF8}{gbsn}腹部饱胀不适、肝区疼痛，肝肿大、压痛及肝功能损害等，部分病例出现黄疸。\end{CJK}  \\
             & \begin{CJK}{UTF8}{gbsn}{\color{blue}原问题:我体检时检查出戊肝，但是我平时生活挺规律的，想要知道戊肝出现的原因有哪些呢？}\end{CJK}\\
    \midrule
    Question & \begin{CJK}{UTF8}{gbsn}{\color{blue}3岁宝宝把整个水果糖咽了，怎么才能知道是咽下去了呢？}\end{CJK}  \\
    Answer   & \begin{CJK}{UTF8}{gbsn}只要是咽后宝宝没有憋气的现象，那就是咽下去了。\end{CJK}  \\
    Top 1    & \begin{CJK}{UTF8}{gbsn}就目前的这种情况首先要确认一下是否已经吞下，一般的情况下宝宝都会有感觉，比如腹疼了，呕吐了等。\end{CJK}  \\
             & \begin{CJK}{UTF8}{gbsn}{\color{blue}原问题:13个月宝宝，昨天发现窗帘上的小挂钩少了一个，怀疑让宝宝误吞了，需要到医院做什么检查吗？}\end{CJK}\\
    Top 2    & \begin{CJK}{UTF8}{gbsn}既然能够咽得下去应该是没事的，你可以注意观察孩子的呼吸状况和面色情况。如有异常问题立即就诊。\end{CJK}  \\
             & \begin{CJK}{UTF8}{gbsn}{\color{blue}原问题:一岁八个月宝宝吃果冻噎住又咽下去了，刚才又喝了点水，有事没有？}\end{CJK}\\
    Top 3    & \begin{CJK}{UTF8}{gbsn}看核的大小，一般都可以排出来，可以密切观察孩子进食情况，只要吃的好，不呕吐，就没问题，如果进\end{CJK}  \\
             & \begin{CJK}{UTF8}{gbsn}食差或呕吐，就要去医院检查了。\end{CJK}  \\
             & \begin{CJK}{UTF8}{gbsn}{\color{blue}原问题:十个月的宝宝吞下荔枝核有没有事，急求答案}\end{CJK}\\
    \bottomrule
    \end{tabular}
    \caption{Examples of retrieval results of DPR model on questions from consultant records}
    \label{tab:retrieval_example_consultant}
\end{table*}

We selected DPR for the case study since it has the best overall performance as the retrieval model.
Table \ref{tab:retrieval_example_wiki},\ref{tab:retrieval_example_KB},\ref{tab:retrieval_example_consultant} shows the  retrieved results using  DPR.
Interestingly, the top-ranked answers are relevant and generally valid, especially for the first case in online consultant records in table \ref{tab:retrieval_example_consultant}  since the number of QA is large and many of them might be redundant and it might lead to \textit{false negatives}. Therefore, although the retrieval metrics (e.g. recall \@5) are relatively low, its retrieval quality is moderately satisfied.  

\end{document}